\pdfoutput=1
\documentclass[final]{cvpr}

\usepackage{times}
\usepackage{epsfig}
\usepackage{graphicx}
\usepackage{amsmath}
\usepackage{amssymb}
\usepackage{tabularx}
\usepackage{booktabs}

\usepackage[pagebackref=true,breaklinks=true,colorlinks,bookmarks=false]{hyperref}

\usepackage{amsfonts}
\usepackage{multirow}
\usepackage{makecell}
\usepackage{array}
\usepackage{bm}
\usepackage{xspace}
\usepackage[title]{appendix}

\newcolumntype{C}[1]{>{\centering\arraybackslash}m{#1}}
\newcolumntype{R}[1]{>{\raggedleft\arraybackslash}m{#1}}
\newcolumntype{P}[1]{>{\raggedright\arraybackslash}p{#1}}
\newcolumntype{M}[1]{>{\centering\arraybackslash}m{#1}}

\def\cvprPaperID{3008} % *** Enter the CVPR Paper ID here
\def\confYear{CVPR 2021}

\title{Unsupervised Multi-Source Domain Adaptation for Person Re-Identification}

\author{
\textbf{Zechen Bai}\textsuperscript{1}, \textbf{Zhigang Wang}\textsuperscript{1}, \textbf{Jian Wang}\textsuperscript{1}, \textbf{Di Hu}$^{2,3\ast}$, \textbf{Errui Ding}$^{1\ast}$\\

\textsuperscript{1}Department of Computer Vision Technology (VIS), Baidu Inc., China,\\
\textsuperscript{2}Gaoling School of Artificial Intelligence, Renmin University of China, Beijing 100872, China, \\
\textsuperscript{3}Beijing Key Laboratory of Big Data Management and Analysis Methods\\

{\tt zechenbai.baidu@outlook.com, dihu@ruc.edu.cn,}\\ 
{\tt \{wangzhigang05,wangjian33,dingerrui\}@baidu.com}
}

\begin{document}

\maketitle

\thispagestyle{empty}

%%%%%%%%% ABSTRACT
\begin{abstract}
Unsupervised domain adaptation (UDA) methods for person re-identification (re-ID) aim at transferring re-ID knowledge from labeled source data to unlabeled target data. Although achieving great success, most of them only use limited data from a single-source domain for model pre-training, making the rich labeled data insufficiently exploited. To make full use of the valuable labeled data, we introduce the multi-source concept into UDA person re-ID field, where multiple source datasets are used during training. However, because of domain gaps, simply combining different datasets only brings limited improvement. In this paper, we try to address this problem from two perspectives, \ie{} domain-specific view and domain-fusion view. Two constructive modules are proposed, and they are compatible with each other. First, a rectification domain-specific batch normalization (RDSBN) module is explored to simultaneously reduce domain-specific characteristics and increase the distinctiveness of person features. Second, a graph convolutional network (GCN) based multi-domain information fusion (MDIF) module is developed, which minimizes domain distances by fusing features of different domains. The proposed method outperforms state-of-the-art UDA person re-ID methods by a large margin, and \textbf{even achieves comparable performance to the supervised approaches without any post-processing techniques}.
\end{abstract}

%%%%%%%%% BODY TEXT
\section{Introduction} \label{intro}
Person{\let\thefootnote\relax\footnote{{$^{*}$Co-Corresponding Auther.}}} re-identification (re-ID) aims at retrieving images of a specified person across different cameras. Recently, supervised person re-ID methods \cite{PCB, MGN, RGA} have made impressive progress. However, when testing on unseen datasets, they usually suffer from dramatic performance degradation. Collecting and annotating enough training data for a specified scenario is expected, but it is labor-intensive and expensive. Therefore, unsupervised domain adaptation (UDA) for person re-ID has attracted an increasing attention. 

Numerous UDA person re-ID methods have been proposed. Among them, the pseudo-label-based branch \cite{udatp, asymmetric, mmt} dominates the state-of-the-art methods. They usually consists of two steps: (1) obtaining a pre-trained model by supervised training on the source domain; (2) training on the target domain by pseudo-label prediction and fine-tune iteratively. Although pseudo-label-based methods are proved effective, most of them only use limited data from a single-source domain for model pre-training, making the rich labeled data insufficiently exploited. This is undoubtedly a great waste of resources.

To make full use of the existing abundant labeled data, we first introduce the multi-source concept into UDA person re-ID field, where multiple source datasets are used in both model pre-training and fine-tuning stages. For the latter, the ground-truth labels of source domain and pseudo labels of target domain together provide supervisions. However, we find simply combining different datasets brings limited improvement or even negative effect. This is because different datasets usually have different characteristics such as color, illumination, camera views, \etc{} This problem is widely known as domain gaps.

In this paper, we try to address this problem from two perspectives, \ie{} domain-specific view and domain-fusion view. As for the former, several successful UDA methods \cite{DSN, CAN, dsbn} use specific network components to capture and eliminate incompatible characteristics of different domains. Nevertheless, most of these methods are originally designed for a close-set problem, \ie{} general classification which has the following properties. (1) Multiple domains share all or part of the label set. (2) Samples have relatively small intra-class distances and large inter-class distances. Due to these two reasons, it is easy for these methods to correctly aggregate intra-class samples even they come from different domains. On the contrary, person re-ID is a fine-grained open-set problem where person images present subtle differences and identities/classes are completely disjoint. Directly applying existing domain-specific methods to multi-domain person re-ID task may improperly aggregate identities from different domains. To tackle this problem, a rectification domain-specific batch normalization (RDSBN) module is proposed in this paper. Inspired by \cite{dsbn}, we use individual BN branches to capture and reduce domain-specific information. Moreover, we exploit a rectification procedure that adaptively tunes BN parameters for each instance to enhance identity-related information and make features more discriminative.

As for the domain-fusion view, some works \cite{DAN, CMD, RTN, JAN} attempt to minimize domain distances by fusing feature distributions. Unfortunately, they also depend on the close-set setting. For re-ID problem, we develop a more flexible GCN-based multi-domain information fusion (MDIF) module to reduce domain distances. There are two main differences between previous GCN-based re-ID methods and our MDIF module. First, from motivation perspective, previous works \cite{sggnn, phgcn} leverage GCN to enhance feature representation, while our MDIF module aims at reducing domain gaps. Second, from working mechanism perspective, previous works mainly follow the formulation of ordinary GCN \cite{gcn}. While in MDIF module, a domain-agent-node concept is proposed which weighted combines features of the same domain as a global representation. Then, our MDIF module enables information to propagate among domain-agent-nodes and instances to conduct domain fusion. During testing, domain-agent-nodes come from the recorded moving average values like BN, thus no extra computation is needed.

The contributions of this work can be summarized as three-fold. 1) We introduce the multi-source concept into UDA person re-ID field. To the best of our knowledge, this is a pioneering work to study the multi-source UDA problem in the person re-ID community. 2) We propose a rectification domain-specific batch normalization (RDSBN) module which can simultaneously reduce domain-specific information and improve the distinctiveness of person features. 3) We develop a GCN based multi-domain information fusion (MDIF) module to pull different domains close in feature space, and shed new light on the effect of GCN on reducing domain gaps.  

In our experiments, we find the proposed two modules are compatible with each other, and the combination of them can outperform state-of-the-art methods by a large margin.

\section{Related Work}
\subsection{Unsupervised Domain Adaptation for Person Re-ID}
Mainstream unsupervised domain adaptation methods for person re-ID can be categorized into two branches. The first branch employs generative adversarial networks (GANs) to transfer the style of labeled images from a source domain to the target domain, and uses transferred images for training. Based on this pipeline, SPGAN \cite{spgan, espgan} propose to preserve similarity of images before and after image translation. Similarly, PTGAN \cite{msmt} leverages semantic segmentation to constrain the consistency of human body regions during style transfer. Unfortunately, the performances of these methods are not satisfactory enough.

On the contrary, the pseudo-label-based branch dominates the state-of-the-art methods in recent years \cite{pul,udatp,ssg,past,asymmetric,adcluster,mmt,dgnet_plus,lin2020corsscam,lin2020soften}. PUL \cite{pul} iteratively obtains pseudo labels by clustering algorithms on the target domain and fine-tunes the model using generated labels. Song \etal{} \cite{udatp} follow this paradigm and provide more theoretical analysis. SSG \cite{ssg} uses both global and local features, and assign pseudo labels for them separately. PAST \cite{past} progressively selects more reliable samples into training set. ACT \cite{asymmetric} designs an asymmetric co-teaching framework to resist label noise generated by clustering algorithms. MMT \cite{mmt} utilizes both hard and sort labels via a mutual mean-teaching framework. DG-Net++ \cite{dgnet_plus} extracts and focuses on identity-related features by a disentangling module. Although these works have achieved great successes, they use only limited data from a single-source domain for model pre-training, making rich labeled data insufficiently exploited. 

\subsection{Multi-Source Domain Adaptation}
Recent years, more multi-source domain adaptation methods have been studied for practical applications. Ganin \etal{} \cite{UDAB} tries to solve domain adaptation problem by adversarial learning. CMSS \cite{CMSS} learns a dynamic curriculum to decide which domains are best for aligning to the target. Peng \etal{} \cite{MMMSDA} aims to transfer knowledge by dynamically aligning moments of multi-domain feature distributions. DSBN \cite{dsbn} narrows domain gaps by incorporating different BN layers for different domains. However, most of them assume there are label overlappings among multiple domains, making them unsuitable for the open-set person re-ID task.

\begin{figure*}
    \centering
    \includegraphics[width=0.9\linewidth]{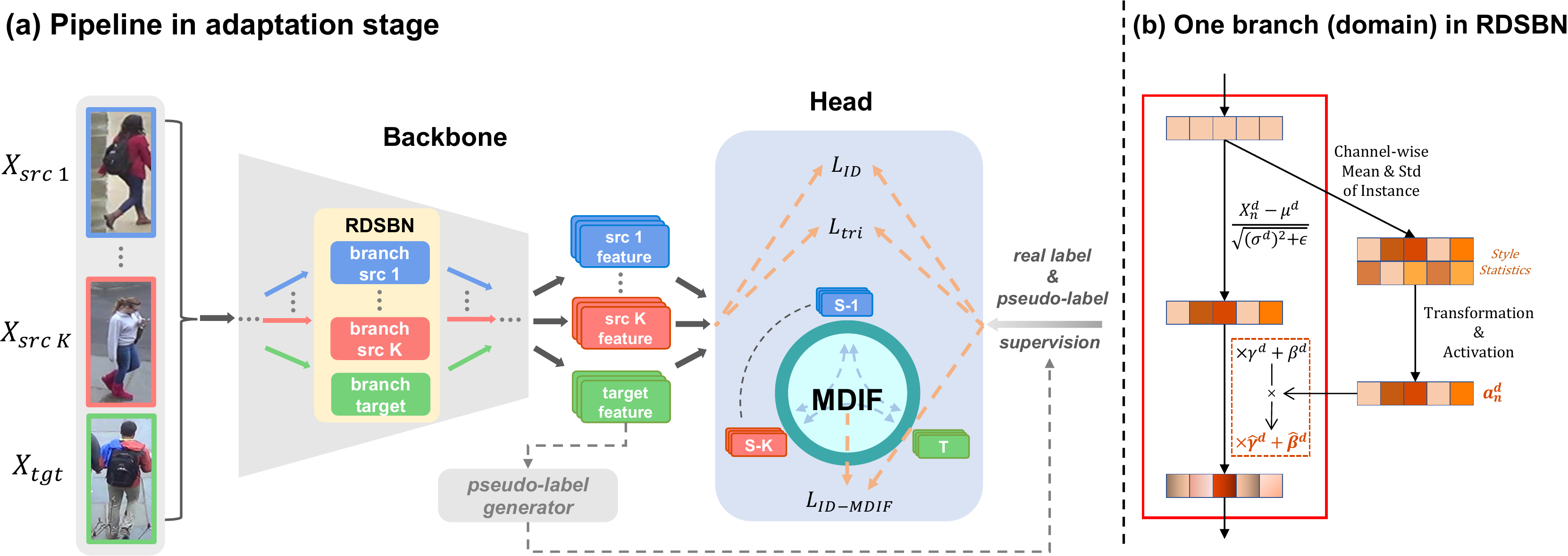}
    \caption{\textbf{(a)} The illustration of the proposed framework, including pseudo-label generator, backbone equipped with RDSBN and head equipped with MDIF. \textbf{(b)} The rectification operation in RDSBN. Best viewed in color.}
    \label{fig:pipeline}
\end{figure*}

\subsection{Feature Normalization}
Recently, simultaneously fusing multiple normalization techniques becomes popular to boost generalization of netwroks. IBN-Net \cite{IBN-Net} manually designs fusion policy for batch normalization (BN) and instance normalization (IN). BIN \cite{BIN} learns adaptive gate weights to assign different importances to BN and IN. Switchable Normalization (SN) \cite{SN} extends this strategy to multiple normalization techniques. Whereas, when applying these methods to UDA person re-ID task, the performance is still unsatisfactory.

\subsection{Graph Convolutional Network Related Methods}
Several works \cite{gcn,gat} concentrate on designing better GCN architectures to address graph-structured problems. Other works extend GCN to different applications, \eg{} action recognition \cite{YanXL18}, anomaly detection \cite{adgcn}, recommendation system \cite{reco} and supervised person re-ID \cite{sggnn}. Great successes are witnessed in these methods, while few works exploit the effect of GCN on reducing domain gaps.

\section{Methodology}

\subsection{Overview}
Under the context of multi-source unsupervised domain adaptive re-ID, we have \textit{K} fully labeled source datasets $\mathcal{S=}$\{\{$Z^{1}, Y^{1}$\},\{$Z^{2}, Y^{2}$\}, ... ,\{$Z^{K}, Y^{K}$\}\}. $Z^{k}$ and $Y^{k}$ denote data samples and ground-truth labels in the \textit{k}-th source domain, respectively. In addition, we have an unlabeled target dataset $\mathcal{T=}$\{$Z^{K+1}$\}. Our goal is to leverage labeled samples in $\mathcal{S}$ and unlabeled samples in $\mathcal{T}$ to learn a re-ID model that generalizes well in the target domain. Note that dataset and domain can be used interchangeably in this paper.

To tackle the aforementioned problem, we propose an unsupervised multi-source domain adaptation framework for person re-ID. As shown in Fig.~\ref{fig:pipeline} (a), the proposed framework can be decoupled into three parts, \ie{} pseudo-label generator, backbone and head. Following most methods \cite{udatp, asymmetric}, the clustering algorithm DBSCAN \cite{dbscan} is employed as our pseudo-label generator which takes all target features as input and outputs clusters as pseudo-labels before each epoch. The backbone is responsible for extracting domain-invariant features from images. It can be chosen from various popular networks. Afterwards, the regular BN layers are replaced by our RDSBN module that mitigates domain-specific characteristics and enhances distinctiveness of person features. Given extracted features, the network head with MDIF module is designed to perform information fusion among multiple domains to further reduce domain gaps. 
The details of RDSBN module, MDIF module and optimization of the proposed framework are described as follows.

\subsection{Rectification Domain-Specific Batch Normalization} \label{RDSBN}
\subsubsection{Batch Normalization Revisit}
A BN layer normalizes features within a mini-batch and performs affine transformation using scale parameter $\bm{\gamma} \in \mathbb{R}^C$ and bias parameter $\bm{\beta} \in \mathbb{R}^C$. Let $\bm{X}_n \in \mathbb{R}^{C \times HW}$ denote the feature map of the \textit{n}-th sample, where $C$, $H$ and $W$ indicate the number of channels, height and width, respectively. BN is formulated as
\begin{equation}
    \mathbf{BN}(\bm{X}_n;\bm{\gamma},\bm{\beta}) = \bm{\gamma} \cdot \frac{\bm{X}_n-\bm{\mu}}{\sqrt{\bm{\sigma}^{2}+\epsilon}} + \bm{\beta},
\end{equation}
where mean value $\bm{\mu} \in \mathbb{R}^C$ and standard deviation $\bm{\sigma} \in \mathbb{R}^C$ are calculated with respect to a mini-batch, $\epsilon$ is a small constant avoiding divide-by-zero, $(\cdot)$ and $\sqrt{\cdot}$ are both channel-wise operations here.

Actually, during training, $\bm{\mu}$ and $\bm{\sigma}$ are estimated through moving average operation with the update momentum $\alpha$. Formally, we donote them by $\bar{\bm{\mu}}$ and $\bar{\bm{\sigma}}$. Given the \textit{t}-th mini-batch, corresponding mean $\bar{\bm{\mu}}^{t}$ and standard deviation $\bar{\bm{\sigma}}^{t}$ are calculated by 
\begin{equation}
    \label{formula:update_mu}
    \bar{\bm{\mu}}^{t} = (1-\alpha)\bar{\bm{\mu}}^{t-1} + \alpha\bm{\mu}^{t},
\end{equation}
\begin{equation}
\label{formula:update_sigma}
    (\bar{\bm{\sigma}}^{t})^2 = (1-\alpha)(\bar{\bm{\sigma}}^{t-1})^2 + \alpha(\bm{\sigma}^{t})^2.
\end{equation}

The final $\bar{\bm{\mu}}$ and $\bar{\bm{\sigma}}$ are used for normalization in the testing stage.

\subsubsection{Design of RDSBN}
The regular BN is performed upon the whole mini-batch. However, sharing parameters across multiple domains is inappropriate due to the existence of domain gap \cite{dsbn}. Inspired by \cite{dsbn}, the proposed RDSBN module also normalizes each domain separately using individual BN branches. Differently, we design a rectification procedure as shown in Fig.~\ref{fig:pipeline} (b). Let $\bm{X}_n^d \in \mathbb{R}^{C \times HW}$ denote a feature map of the \textit{n}-th sample in \textit{d}-th domain, RDSBN module can be written as

\begin{equation}
    \mathbf{RDSBN}(\bm{X}_n^d;\bm{\gamma}^d,\bm{\beta}^d, \bm{a}_n^d) = \bm{a}_n^d \cdot \bm{\gamma}^d \cdot \frac{\bm{X}_n^d-\bm{\mu}^d}{\sqrt{({\bm{\sigma}^d})^2+\epsilon}} + \bm{a}_n^d \cdot \bm{\beta}^d.
\end{equation}
For simplicity, the moving average version is not introduced again. The superscript $d$ indicates that calculations are carried out within a specific domain.  $\bm{a}_n^d \in \mathbb{R}^C$ is a channel-wise weight that rectifies $\bm{\gamma}^d$ and $\bm{\beta}^d$ to rearrange features. This is expected to explicitly enhance the identity-related style information. In particular, inspired by \cite{adain}, the style-related statistics, \ie{} channel-wise mean $\bm{\mu}_n^d$ and standard deviation $\bm{\sigma}_n^d$ of each instance are adopted to estimate $\bm{a}_n^d$.
\begin{equation}
    \bm{a}_n^d = Sigmoid(\mathbf{1}^M * (\bm{r}^d * [\bm{\mu}_n^d, \bm{\sigma}_n^d])), 
\end{equation}
where $Sigmoid(\cdot)$ means regular sigmoid activation function, $*$ indicates matrix multiplication, $\mathbf{1}^M \in \mathbb{R}^{1 \times M}$ is an all-one matrix for dimension reduction, $\bm{r}^d \in \mathbb{R}^{M \times 2}$ is a learnable parameter, [$\cdot$] denotes concatenate operation that combines $\bm{\mu}_n^d$ and $\bm{\sigma}_n^d$ into a two-dimension matrix, \ie{} $[\bm{\mu}_n^d, \bm{\sigma}_n^d] \in \mathbb{R}^{2 \times C}$. 

The proposed rectification procedure is simple yet very effective, which plays an important role in increasing the distinctiveness of person features after reducing domain gaps. This is critical for the fine-grained open-set person re-ID problem. Experiments show the superiority of our RDSBN module. See section \ref{RDSBN_exp} for details.

\begin{figure}
    \centering
    \includegraphics[width=0.93\linewidth]{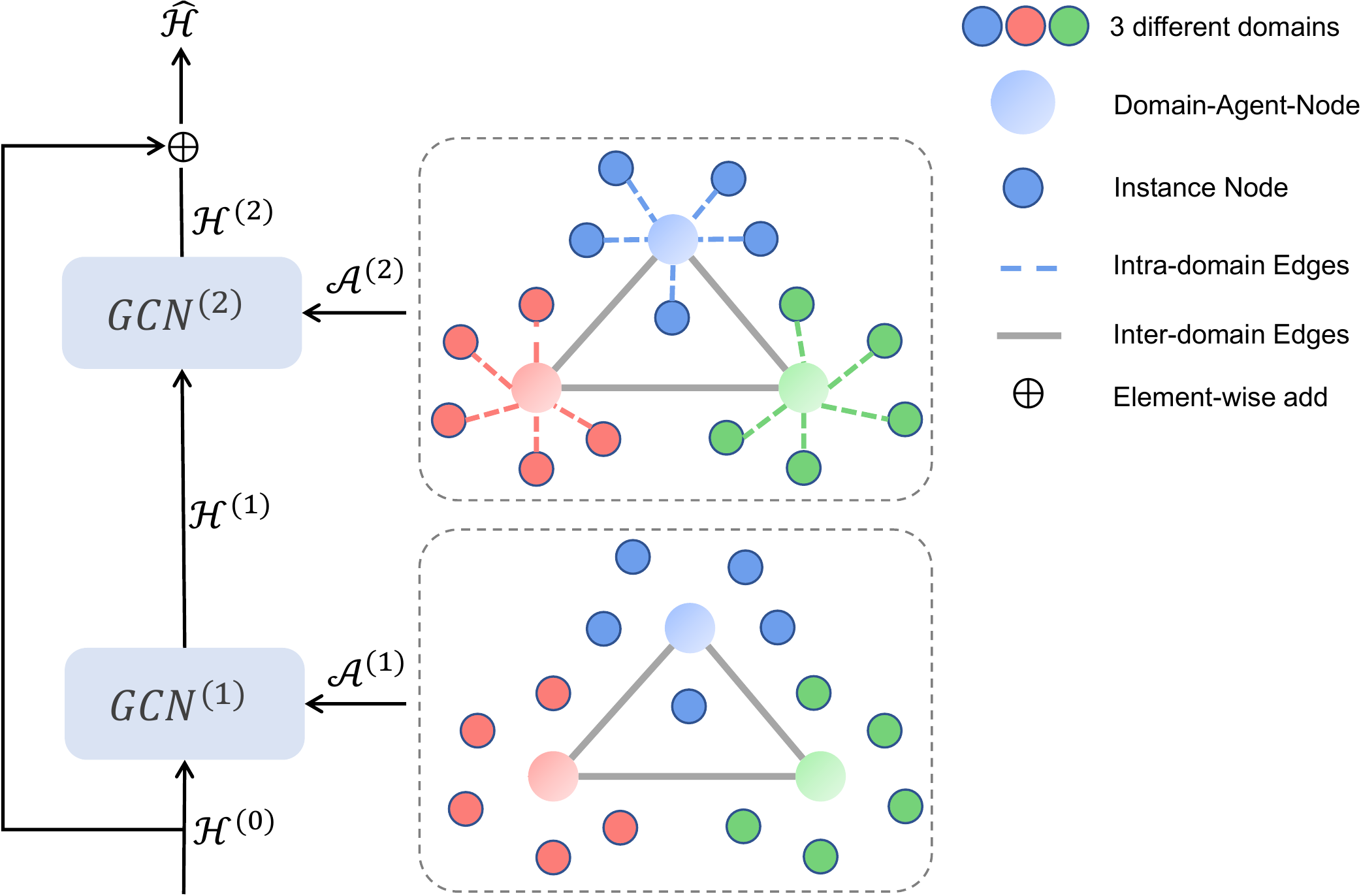}
    \caption{The illustration of GCN-based Multi-Domain Information Fusion module. For convenience, we present two source domains and one target domain here. Best viewed in color.}
    \label{fig:gcn}
\end{figure}

\subsection{Multi-Domain information Fusion}
\subsubsection{Domain-Agent-Node}
We also attempt to fuse domain information by GCN to reduce domain distances. One naive way to apply GCN here is building a graph that connects instances from different domains. However, directly fusing instance-level features is prone to incur identity-related noises. Therefore, we propose a domain-agent-node as the global domain representation. It is obtained by weighted combining features of the same domain. Formally, the domain-agent-node is formulated as below

\begin{equation}
    \bm{agt}^d = \sum_{n=1}^Q w_n^d \cdot \bm{X}_n^d, 
\end{equation}
where $\bm{agt}^d$ denotes the agent-node of the \textit{d}-th domain, $Q$ means the number of samples belonging to the same domain within a mini-batch, $w_n^d = f(\bm{X}_n^d)$ is a learnable scalar weight and normalized by $w_n^d = w_n^d / \sum_{m=1}^Q w_m^d$. $f(\cdot)$ indicates a fully connected layer.

In order to estimate the domain-agent-node more stably, we adopt the moving average technique during training like BN
\begin{equation}
    (\bar{\bm{agt}^d})^{t} = (1-\alpha)(\bar{\bm{agt}^d})^{t-1} + \alpha(\bar{\bm{agt}^d})^{t}, 
\end{equation}
where $t$ and $\alpha$ have the same meanings with Eq. \ref{formula:update_mu} and Eq. \ref{formula:update_sigma}. Similar to BN, the recorded $\bar{\bm{agt}^d}$ is used for inference in the testing stage.

\subsubsection{Graph construction}
Let $G(V, E)$ denote a graph where $V$ is the vertex (node) set and $E$ is the edge set. As shown in Fig.~\ref{fig:gcn}, we stack two GCN layers to construct $G(V, E)$, in which the first layer is responsible for fusing global domain representations, and the second layer ensures all instances are capable of receiving information from other domains. Without loss of generality, assume there are 3 domains in a mini-batch, each domain contains $Q$ samples and 1 domain-agent-node. Thus there are totally $3\times(Q+1)$ nodes. We denote $\mathcal{H}^{(l)} \in \mathbb{R}^{(3Q+3) \times C}$ as node features where the superscript $(l)$ indicates the \textit{l}-th GCN layer. As for the edge set $E$, it can be represented by adjacency matrixes $\mathcal{A}^{(l)} \in \mathbb{R}^{(3Q+3) \times (3Q+3)}$ of two GCN layers. For inter-domain edges, the connections only exist among 3 domain-agent-nodes in both $\mathcal{A}^{(1)}$ and $\mathcal{A}^{(2)}$. For intra-domain edges, each instance is connected to its affiliated domain-agent-node in $\mathcal{A}^{(2)}$. Note that the intra-connections are omitted in $\mathcal{A}^{(1)}$, since in the first layer of GCN, domain-agent-nodes have not incorporated information from each other and cannot provide fused features for corresponding intra-domain nodes.

Based on the above connection relationship, $\mathcal{A}^{(l)}$ can be formulated as
\begin{equation}
    A^{(l)}_{i,j}=\begin{cases}1 & i=j \; or \; node \;i \; connects \; node \; j \\ 0 & otherwise \end{cases} , 
\end{equation}
where $i$ and $j$ represent row index and column index of $\mathcal{A}^{(l)}$, respectively. $\mathcal{A}^{(l)}$ is then normalized by its degree matrix $D^{(l)}$
\begin{equation}
    \tilde{\mathcal{A}^{(l)}} = (D^{(l)})^{-\frac{1}{2}}\mathcal{A}^{(l)}(D^{(l)})^{-\frac{1}{2}},
\end{equation}
where $D^{(l)}_{i, i}=\sum_j \mathcal{A}^{(l)}_{i, j}$. Afterwards, each GCN layer can be written as a non-linear transformation
\begin{equation}
    \mathcal{H}^{(l)}=\rho(\tilde{\mathcal{A}}^{(l)} \mathcal{H}^{(l-1)} \mathcal{W}^{(l)}),
\end{equation}
where $\mathcal{W}^{(l)} \in \mathbb{R}^{C \times C}$ is a learnable transformation matrix and $\rho$ denotes a non-linear function that is LeakyReLU \cite{leakyrelu} in our method. Let $\mathcal{H}^{(0)} \in \mathbb{R}^{3Q \times C}$ denote the input features to our MDIF module, the corresponding $\mathcal{H}^{(2)}$ without domain-agent-node features is output and treated as residuals to be added to original features
\begin{equation}
    \hat{\mathcal{H}} = \mathcal{H}^{(0)} + \mathcal{H}^{(2)}.
\end{equation}
Finally, we use the fused features $\hat{\mathcal{H}}$ as sample representations. For more detailed explanation of the MDIF inference procedure, please refer to \textbf{Supplementary}.

\begin{table*}[t]
 	\centering
 	\caption{
	Ablation studies on different number of source domains with Market1501 as target domain.}
	\vspace{4pt}
	\begin{tabular}{P{1.5cm}|C{0.75cm}C{0.75cm}C{0.75cm}C{0.75cm}|C{0.75cm}C{0.75cm}C{0.75cm}C{0.75cm}|C{0.8cm}C{0.8cm}C{0.8cm}C{0.85cm}}
 	\Xhline{2\arrayrulewidth}
	\multicolumn{1}{c|}{\multirow{2}{*}{Methods}} & \multicolumn{4}{c|}{1-source: Duke} & \multicolumn{4}{c|}{2-source: Duke+CUHK} & \multicolumn{4}{c}{3-source: Duke+CUHK+MSMT}\\
	\cline{2-13}
	\multicolumn{1}{c|}{} & mAP & R-1 & R-5 & R-10 & mAP & R-1 & R-5 & R-10 & mAP & R-1 & R-5 & R-10 \\
    \Xhline{2\arrayrulewidth}
   \multicolumn{1}{l|}{Direct Transfer} & 27.7 & 56.2 & 71.8 & 76.6 & 36.0 & 63.5 & 78.5 & 83.3 & 40.3 & 67.5 & 82.0 & 86.4 \\
   \multicolumn{1}{l|}{MMT (DBSCAN)} & 74.6 & 88.4 & 96.2 & 97.8 & 75.3 & 89.5 & 96.6 & 98.0 & 74.8 & 89.3 & 96.2 & 97.7 \\
   \multicolumn{1}{l|}{MMT-with-Source} & 68.9 & 87.1 & 95.0 & 96.7 & 75.3 & 89.6 & 96.5 & 97.6 & 77.7 & 90.8 & 96.9 & 97.9\\
   \multicolumn{1}{l|}{MMT + Ours} & \textbf{81.5} & \textbf{92.9} & \textbf{97.6} & \textbf{98.4} & \textbf{85.2} & \textbf{94.2} & \textbf{98.0} & \textbf{98.8} & \textbf{86.0} & \textbf{94.8} & \textbf{97.9} & \textbf{98.6}\\
   \Xhline{2\arrayrulewidth}
	\end{tabular}\\
\label{tab:multi-source}
\end{table*}

\subsection{Optimization}
\subsubsection{Two-Stage Training}
Following the popular pipeline, we adopt a two-stage training scheme including source-model pre-training and domain adaptation fine-tuning. 

For the first stage, a person re-ID model with DSBN layers is trained by optimizing ID loss (cross-entropy loss) and triplet loss \cite{triplet}

\begin{equation}
    \mathcal{L}_{src} = \mathcal{L}_{ID} + \mathcal{L}_{tri}.
\end{equation}
Duo to the limitation of page-length and generality of these two losses, we omit their detailed formulation. Please refer to corresponding references for details.

For the second stage, the re-ID model is further equipped with RDSBN and MDIF modules: replacing DSBN with RDSBN and stacking MDIF module on top of the backbone as shown in Fig~\ref{fig:pipeline}. Both ground-truth labels and pseudo labels provide supervisions in this stage. We select ID loss to train MDIF module. The original losses building upon the input features of MDIF module are retained to preserve their representativeness of identities. So that in the second stage, the total loss can be written as
\begin{equation}
    \mathcal{L}_{ada} = \mathcal{L}_{ID} + \mathcal{L}_{ID-MDIF} + \mathcal{L}_{tri}.
\end{equation}

\section{Experiment}
\subsection{Datasets and Evaluation Metrics}
We evaluate the proposed method on four widely-used person re-ID datasets: Market1501\cite{market1501}, DukeMTMC-reID\cite{ristani2016MTMC}, CUHK03\cite{cuhk03} and MSMT\cite{msmt}. Mean average precision (mAP) and CMC rank-1, rank-5, rank-10 accuracies are adopted to evaluate the performances.

\subsection{Implementation Details}
% Our method is designed to be optimized by two stages: . 
In both source-model pre-training and domain adaptation fine-tuning stages, ResNet50 \cite{resnet} is adopted as the backbone network. Experiments are carried out on two NVIDIA-P100 GPUs. We follow the experimental settings of \cite{mmt}, details are listed as below. Adam optimizer \cite{adam} is utilized to optimize networks with a weight decay of 0.0005. The input image is uniformly resized to $256 \times 128$. Several data augmentation techniques are performed such as random flipping and random crop. Random erasing is only used in fine-tuning stage.

\textbf{Stage1: source-model pre-training.} In this stage, the ResNet50-DSBN network is initialized with ResNet50 weights pre-trained on ImageNet. Each training mini-batch contains 32 person images of 8 identities (4 for each identity). The learning rate is initialized as 0.00035 and is decreased to 1/10 of its previous value on the 40-th and 70-th epoch in total 80 epochs.

\textbf{Stage2: domain adaptation fine-tuning.} The proposed network ResNet50-RDSBN-MDIF is initialized with weights of the above ResNet50-DSBN. For each training mini-batch, we select $32\times(K+1)$ person images (8 identities in each domain, $K$ source datasets and 1 target dataset). The learning rate is fixed to 0.00035 for overall 40 training epochs. During testing, we only use the target domain RDSBN branch for feature extraction and fetch the fused features after MDIF module for inference.

\begin{table}[htb]
	\centering
	\caption{Ablation studies on RDSBN module. Due to space limitation, we only show a portion of the results here and more comparisons can be found in \textbf{Supplementary}.}
	\vspace{4pt}
	\begin{tabular}{P{4.0cm}|C{0.55cm}C{0.55cm}C{0.55cm}C{0.73cm}}
 	\Xhline{2\arrayrulewidth}
	\multicolumn{1}{c|}{\multirow{2}{*}{Methods}} & \multicolumn{4}{c}{Duke+CUHK$\to$Market}  \\
	\cline{2-5}
	\multicolumn{1}{c|}{} & mAP & R-1 & R-5 & R-10 \\ 
    % \hline
    \Xhline{2\arrayrulewidth}
    ResNet50-\textit{BN} (baseline) & 70.1 & 86.7 & 95.0 & 96.5 \\
    ResNet50-\textit{BIN} & 70.5 & 87.8 & 94.9 & 96.5  \\
    ResNet50-\textit{IBN} & 70.7 & 88.0 & 95.3 & 97.3 \\
    ResNet50-\textit{SN} & 65.8 & 84.8 & 93.6 & 95.8 \\
    ResNet50-\textit{DSBN} & 74.1 & 89.2 & 96.2 & \textbf{97.9}   \\
    ResNet50-\textit{RDSBN} & \textbf{78.2} & \textbf{91.0} & \textbf{96.8} & \textbf{97.9}  \\
    \Xhline{2\arrayrulewidth}
	\end{tabular}
\label{tab:rdsbn}
\end{table}

\subsection{Ablation Studies}

\subsubsection{Necessity of Multiple Source Domains}
A key problem to this work is the necessity of introducing multiple source domains. Experiments results with respect to the number of source datasets are shown in Table~\ref{tab:multi-source}. Along the `Methods' column, `Direct Transfer' means directly using the pre-trained source-model for testing. `MMT (DBSCAN)' follows the setting of \cite{mmt} where one or more source datasets are only used for pre-training. `MMT-with-Source' further brings source data into fine-tuning stage. We observe that increasing the number of source domains brings stable performance improvements in `Direct Transfer'. It indicates that introducing more source data is beneficial to the generalization ability of a re-ID model. However, simply combining multiple source datasets (`MMT (DBSCAN)' and `MMT-with-Source') brings only limited improvements or even negative effect. In comparison, our method can boost performance significantly when using more datasets. In a nutshell, introducing multiple source datasets is beneficial, and our method can effectively leverage these data.

\subsubsection{Effectiveness of RDSBN}
\label{RDSBN_exp}
To evaluate the effectiveness of our RDSBN module, we compare it with the most related DSBN \cite{dsbn} and several powerful normalization techniques in Table~\ref{tab:rdsbn}. For simplicity, two source datasets and one target dataset are used for verification. All methods in this experiment are built upon the most ordinary pseudo-label-based UDA pipeline (Sec.~\ref{intro}), and all of them use source data in domain adaptation fine-tuning stage for fair comparison. We first create a baseline model equipped with regular BN, which has already achieved a good result. BIN \cite{BIN} and IBN \cite{IBN-Net} further boost the performance due to their abilities of increasing generalization. However, the improvement is not significant, SN \cite{SN} even causes performance degradation. DSBN outperforms the above methods, which shows advantage of domain-specific design in dealing with multi-domain problem. However, DSBN is not designed for the fine-grained open-set re-ID problem. In comparison, our RDSBN module considerably improves the results, which outperforms DSBN by 4.1\% mAP and 1.8\% top-1 accuracy. This result demonstrates the importance and effectiveness of the proposed rectification procedure.

\begin{table}[htb]
	\centering
	\caption{Ablation studies on MDIF module. See \textbf{Supplementary} for more results.}
	\vspace{4pt}
	\begin{tabular}{P{4.0cm}|C{0.55cm}C{0.55cm}C{0.55cm}C{0.73cm}}
 	\Xhline{2\arrayrulewidth}
	\multicolumn{1}{c|}{\multirow{2}{*}{Methods}} & \multicolumn{4}{c}{Duke+CUHK$\to$Market} \\
	\cline{2-5}
	\multicolumn{1}{c|}{} & mAP & R-1 & R-5 & R-10 \\
% 	\hline \hline
    \Xhline{2\arrayrulewidth}
    ResNet50-\textit{BN} (baseline) & 70.1 & 86.7 & 95.0 & 96.5  \\
    ResNet50-\textit{BN-MDIF} & 73.2 & 88.4 & 95.2 & 97.1 \\
    \hline
    ResNet50-\textit{DSBN} & 74.1 & 89.2 & 96.2 & 97.9   \\
    ResNet50-\textit{DSBN-MDIF} & 78.3 & 91.3 & 96.7 & 97.9   \\
    \hline
    ResNet50-\textit{RDSBN} & 78.2 & 91.0 & 96.8 & 97.9  \\
    ResNet50-\textit{RDSBN-MDIF} & 79.4 & 92.1 & 97.1 & 98.1  \\
    \Xhline{2\arrayrulewidth}
	\end{tabular}
\label{tab:mdif_ablation}
\end{table}

\begin{table}[h]
	\centering
	\caption{Analysis of MDIF module design. See \textbf{Supplementary} for more results.}
	\vspace{4pt}
	\begin{tabular}{P{4.0cm}|C{0.55cm}C{0.55cm}C{0.55cm}C{0.73cm}}
 	\Xhline{2\arrayrulewidth}
	\multicolumn{1}{c|}{\multirow{2}{*}{Methods}} & \multicolumn{4}{c}{Duke+CUHK$\to$Market} \\
	\cline{2-5}
	\multicolumn{1}{c|}{} & mAP & R-1 & R-5 & R-10 \\ 
    \Xhline{2\arrayrulewidth}
    baseline & 70.1 & 86.7 & 95.0 & 96.5  \\
    Instance Graph & 72.4 & 87.6 & 94.2 & 96.8  \\
    Mean Agent Node & 72.9 & 88.0 & 94.9 & 96.8  \\
    Weighted Mean Agent Node & \textbf{73.2} & \textbf{88.4} & \textbf{95.2} & \textbf{97.1}  \\
    intra-connections in $A^{(1)}$ & 73.1 & 87.7 & 95.1 & 97.0  \\
    \Xhline{2\arrayrulewidth}
	\end{tabular}
\label{tab:mdif_design}
\end{table}

\begin{figure}
    \centering
    \includegraphics[width=0.9\linewidth]{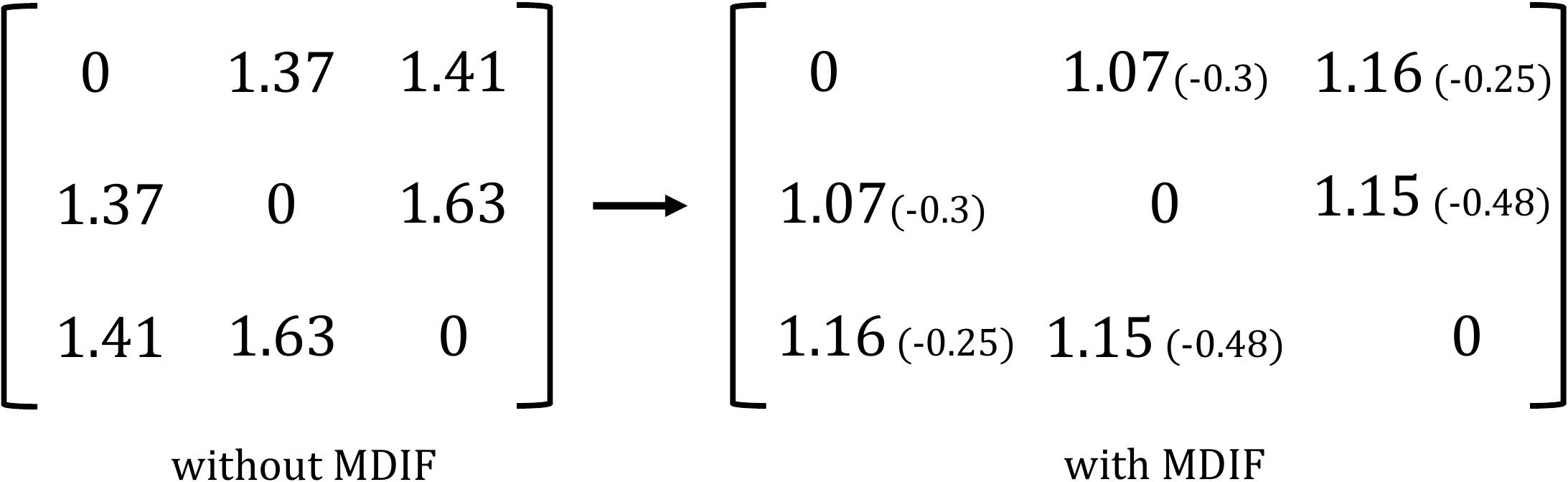}
    \caption{Pair-wise Euclidean distances among three domains: Market, Duke and CUHK03.}
    \label{fig:fusion_distance}
\end{figure}

\begin{table*}[htb]
 	\centering
 	\caption{
	Comparison with state-of-the-art methods about unsupervised domain adaptive re-ID.
	(*) indicates the implementation is based on the authors' code.}
	\vspace{4pt}
% 	\begin{center}
	\centering
	\begin{tabular}{P{1.4cm}C{1.5cm}|C{0.7cm}C{0.7cm}C{0.7cm}C{0.75cm}|C{0.7cm}C{0.7cm}C{0.7cm}C{0.75cm}}
 	\Xhline{2\arrayrulewidth}
	\multicolumn{2}{c|}{\multirow{2}{*}{Methods}} & \multicolumn{4}{c|}{Duke$\to$Market} & \multicolumn{4}{c}{Market$\to$Duke} \\
	\cline{3-10}
	\multicolumn{2}{c|}{} & mAP & R-1 & R-5 & R-10 & mAP & R-1 & R-5 & R-10 \\
    \hline
    % \Xhline{2\arrayrulewidth}
    \multicolumn{3}{l}{\textit{\textbf{Single-Source} UDA re-ID methods:}} \\
    \hline 
    
    \multicolumn{1}{l|}{PAST~\cite{past}} & ICCV'19 & 54.6 & 78.4 & - & - & 54.3 & 72.4 & - & - \\
    \multicolumn{1}{l|}{SSG~\cite{ssg}} & ICCV'19 & 58.3 & 80.0 & 90.0 & 92.4 & 53.4 & 73.0 & 80.6 & 83.2 \\
    \multicolumn{1}{l|}{ECN++~\cite{zhong2020learning}} & TPAMI'20 & 63.8 & 84.1 & 92.8 & 95.4 & 54.4 & 74.0 & 83.7 & 87.4 \\
  \multicolumn{1}{l|}{MMCL~\cite{wang2020unsupervised}} & CVPR'20 & 60.4 & 84.4 & 92.8 & 95.0 & 51.4 & 72.4 & 82.9 & 85.0 \\
  \multicolumn{1}{l|}{SNR~\cite{jin2020style}} & CVPR'20 & 61.7 & 82.8 & - & - & 58.1 & 76.3 & - & - \\
  \multicolumn{1}{l|}{AD-Cluster~\cite{adcluster}} & CVPR'20 & 68.3 & 86.7 & 94.4 & 96.5 & 54.1 & 72.6 & 82.5 & 85.5 \\
  \multicolumn{1}{l|}{DG-Net++~\cite{dgnet_plus}} & ECCV'20 & 61.7 & 82.1 & 90.2 & 92.7 & 63.8 & 78.9 & 87.8 & 90.4 \\
  \multicolumn{1}{l|}{NRMT~\cite{nrmt}} & ECCV'20 & 71.7 & 87.8 & 94.6 & 96.5 & 62.2 & 77.8 & 86.9 & 89.5 \\
  \multicolumn{1}{l|}{MMT~\cite{mmt} (DBSCAN)*} & ICLR'20 & 74.6 & 88.4 & 96.2 & 97.8 & 61.0 & 75.1 & 87.3 & 91.2\\
   \multicolumn{1}{l|}{Baseline} & - & 58.7 & 82.4 & 90.7 & 94.2 & 49.2 & 67.5 & 80.8 & 86.2 \\
   \multicolumn{1}{l|}{Baseline+\textbf{Ours}} & -  & 68.4 & 85.9 & 94.3 & 96.5 & 55.8 & 72.9 & 84.4 & 88.1  \\
   \multicolumn{1}{l|}{MMT+\textbf{Ours}} & -  & \textbf{81.5} & \textbf{92.9} & \textbf{97.6} & \textbf{98.4} & \textbf{66.6} & \textbf{80.3} & \textbf{89.1} & \textbf{92.6} \\
   \hline
   \multicolumn{3}{l}{\textit{\textbf{Multi-Source} UDA methods (source+CUHK03+MSMT):}} \\
    \hline 
    \multicolumn{1}{l|}{MMT+GRL*~\cite{grl}} & ICML'15 & 77.1 & 90.4 & 96.8 & 97.8 & 64.3 & 77.6 & 88.1 & 91.3 \\
    \multicolumn{1}{l|}{MMT+MomentMatching*~\cite{peng2019moment}} & ICCV'19 & 78.1 & 91.2 & 96.8 & 98.0 & 64.2 & 77.6 & 87.9 & 91.4 \\
    \multicolumn{1}{l|}{MMT+DSBN*~\cite{dsbn}} & CVPR'19 & 81.1 & 92.8 & 97.3 & 98.5 & 65.6 & 79.6 & 89.1 & 92.1 \\
	\multicolumn{1}{l|}{MMT+\textbf{Ours}} & - & \textbf{86.0} & \textbf{94.8} & \textbf{97.9} & \textbf{98.6} & \textbf{68.9} & \textbf{82.1} & \textbf{90.4} & \textbf{93.0} \\
 	\hline
 	\multicolumn{3}{l}{\textit{\textbf{Supervised} methods:}} \\
    \hline 
    \multicolumn{1}{l|}{PCB~\cite{PCB}} & ECCV'18 & 81.6 & 93.8 & 97.5 & 98.5 & 69.2 & 83.3 & - & - \\
    \multicolumn{1}{l|}{bag-of-tricks~\cite{bagoftricks}} & CVPRW'19 & 85.9 & 94.5 & 98.2 & 99.0 & 76.4 & 86.4 & 93.9 & 96.1 \\
    \Xhline{2.5\arrayrulewidth}
	\end{tabular}\\
	\vspace{15pt}

	\begin{tabular}{P{1.4cm}C{1.5cm}|C{0.7cm}C{0.7cm}C{0.7cm}C{0.75cm}|C{0.7cm}C{0.7cm}C{0.7cm}C{0.75cm}}
 	\Xhline{2.5\arrayrulewidth}
	\multicolumn{2}{c|}{\multirow{2}{*}{Methods}} & \multicolumn{4}{c|}{Market$\to$MSMT} & \multicolumn{4}{c}{Duke$\to$MSMT} \\
	\cline{3-10}
	\multicolumn{2}{c|}{} & mAP & R-1 & R-5 & R-10 & mAP & R-1 & R-5 & R-10 \\
	\hline
	% \Xhline{2\arrayrulewidth}
	\multicolumn{3}{l}{\textit{\textbf{Single-Source} UDA re-ID methods:}} \\
	\hline
    \multicolumn{1}{l|}{SSG~\cite{ssg}} & ICCV'19 & 13.2 & 31.6 &- & 49.6 & 13.3 & 32.2 & - & 51.2 \\
    \multicolumn{1}{l|}{ECN++~\cite{zhong2020learning}} & TPAMI'20 & 15.2 & 40.4 & 53.1 & 58.7 & 16.0 & 42.5 & 55.9 & 61.5 \\
   \multicolumn{1}{l|}{MMCL~\cite{wang2020unsupervised}} & CVPR'20 & 15.1 & 40.8 & 51.8 & 56.7 & 16.2 & 43.6 & 54.3 & 58.9 \\
   \multicolumn{1}{l|}{DG-Net++~\cite{dgnet_plus}} & ECCV'20 & 22.1 & 48.4 & 60.9 & 66.1 & 22.1 & 48.8 & 60.9 & 65.9 \\
    \multicolumn{1}{l|}{MMT~\cite{mmt} (DBSCAN)*} & ICLR'20 & 25.7 & 51.9 & 65.3 & 70.9 & 28.1 & 56.1 & 68.9 & 74.3 \\
    \multicolumn{1}{l|}{Baseline} & - & 8.1 & 19.9 & 29.6 & 35.0 & 9.3 & 23.4 & 33.5 & 39.4 \\
   \multicolumn{1}{l|}{Baseline+\textbf{Ours}} & - & 20.7 & 46.9 & 60.0 & 65.0 & 21.3 & 47.7 & 60.8 & 66.4 \\
	\multicolumn{1}{l|}{MMT+\textbf{Ours}} & - & 30.9 & 61.2 & 73.1 & 77.4 & 33.6 & 64.0 & 75.6 & 79.6 \\
	\hline
   \multicolumn{3}{l}{\textit{\textbf{Multi-Source} UDA methods (Market+Duke+CUHK03):}} \\
    \hline 
    \multicolumn{1}{l|}{MMT+GRL*~\cite{grl}} & ICML'15 & 22.6 & 46.3 & 59.8 & 65.8 & 22.6 & 46.3 & 59.8 & 65.8  \\
    \multicolumn{1}{l|}{MMT+MomentMatching*~\cite{peng2019moment}} & ICCV'19 & 22.7 & 47.1 & 60.4 & 65.8 & 22.7 & 47.1 & 60.4 & 65.8 \\
    \multicolumn{1}{l|}{MMT+DSBN*~\cite{dsbn}} & CVPR'19 & 22.6 & 49.2 & 62.1 & 67.4 & 22.6 & 49.2 & 62.1 & 67.4  \\
	\multicolumn{1}{l|}{MMT+\textbf{Ours}} & - & \textbf{34.9} & \textbf{64.7} & \textbf{76.2} & \textbf{80.2} & \textbf{34.9} & \textbf{64.7} & \textbf{76.2} & \textbf{80.2} \\
 	%\Xhline{2\arrayrulewidth}
 	\Xhline{2.5\arrayrulewidth}
	\end{tabular}
\label{tab:sota}
% \vspace{-2pt}
\end{table*}

\subsubsection{Effectiveness of MDIF and Structure Analysis}
Performances of the proposed MDIF module under different model settings are reported in Table \ref{tab:mdif_ablation}. We can observe that placing MDIF module on different base models always bring obvious improvements. This shows the effectiveness and robustness of the MDIF module. 

\begin{table*}[!htb]
 	\centering
 	\caption{
	Evaluate domain adaptation model on source domains with market1501 as target dataset.}
	\vspace{4pt}
	\begin{tabular}{P{1.5cm}|C{0.6cm}C{0.6cm}C{0.6cm}C{0.75cm}|C{0.6cm}C{0.6cm}C{0.6cm}C{0.75cm}|C{0.6cm}C{0.6cm}C{0.6cm}C{0.75cm}}
 	\Xhline{2\arrayrulewidth}
	\multicolumn{1}{c|}{\multirow{2}{*}{Methods}} & \multicolumn{4}{c|}{Dukemtmc} & \multicolumn{4}{c|}{CUHK03} & \multicolumn{4}{c}{MSMT}\\
	\cline{2-13}
	\multicolumn{1}{c|}{} & mAP & R-1 & R-5 & R-10 & mAP & R-1 & R-5 & R-10 & mAP & R-1 & R-5 & R-10 \\
    \Xhline{2\arrayrulewidth}
     \multicolumn{1}{l|}{PCB\cite{PCB}} & 69.2 & 83.3 & - & - & 57.5 & 63.7 & - & - & - & - & - & - \\
    \multicolumn{1}{l|}{bag-of-tricks\cite{bagoftricks}} & 76.4 & 86.4 & 93.9 & 96.1 & - & - & - & - & - & - & - & - \\
   \multicolumn{1}{l|}{Pre-trained model} & 65.4 & 79.5 & 89.6 & 92.6 & 61.4 & 64.0 & 77.0 & 87.5 & 33.1 & 60.4 & 74.6 & 79.9 \\
   \multicolumn{1}{l|}{MMT (DBSCAN)} & 16.2 & 30.0 & 43.9 & 51.0 & - & - & - & - & - & - & - & - \\
   \multicolumn{1}{l|}{MMT (3 Sources + 1 target)} & 72.5 & 82.8 & 91.9 & 95.0 & 82.9 & 85.0 & 93.0 & 96.0 & 46.0 & 71.2 & 83.1 & 86.8 \\
   \multicolumn{1}{l|}{Ours (3 Sources + 1 target)} & \textbf{78.1} & \textbf{88.1} & \textbf{94.4} & \textbf{96.3} & \textbf{87.3} & \textbf{90.5} & \textbf{94.0} & \textbf{97.0} & \textbf{54.2} & \textbf{78.3} & \textbf{87.8} & \textbf{90.8} \\
   \Xhline{2\arrayrulewidth}
	\end{tabular}\\
\label{tab:test-on-source}
\end{table*}

Another experiment is conducted to directly validate whether MDIF module can decrease domain distances. We use the average feature of the whole domain as its representation. Pair-wise Euclidean distances among three domains are calculated and shown in Fig. \ref{fig:fusion_distance}. It can be seen that the proposed MDIF module reduces domain distances to a large extent.

The effect of domain-agent-node is also analyzed. As shown in Table \ref{tab:mdif_design}, we compare it with the `Instance Graph' that directly constructs graph based on all instances within a mini-batch. The adjacency matrix of `Instance Graph' is constructed by calculating cosine similarities of instances. Moreover, we explore two variants of the domain-agent-node, \ie{} `Mean Agent Node' and `Weighted Mean Agent Node'. The former directly averages node features , while the latter weighted combines node features. Both of them outperform `Instance Graph', indicating the proposed domain-agent-node can better represent a domain and promote domain information fusion. Finally, the better one, \ie{} `Weighted Mean Agent Node' is selected in our method. Based on `Weighted Mean Agent Node', a detailed experiment is conducted to verify the irrationality of intra-connections in the first GCN layer of MDIF module. The inferior result validates our assumption, namely, domain-agent-nodes have not been fused at this time and should not propagate information to their intra-domain nodes.

\subsection{Comparison with state-of-the-arts}
To further prove the superiority of the proposed method, we compare it with several state-of-the-art methods on four domain adaptation tasks as MMT \cite{mmt}: Market-to-Duke, Duke-to-Market, Market-to-MSMT and Duke-to-MSMT. The results are shown in Table \ref{tab:sota}. For fair comparison, we first conduct single-source domain adaptation experiments. Even though, our method significantly improves the baseline. When combining our method with the state-of-the-art work MMT, it outperforms all comparison UDA person re-ID works by a large margin. Specifically, our method achieves 81.5\% and 66.6\% mAP on Market-to-Duke and Duket-to-Market tasks, which outperforms MMT by 6.9\% and 5.6\%, respectively. This result indicates that our method can bring stable improvements even on the state-of-the-art method.

When extending our method to the multi-source version, another round of performance boost can be seen. Since no previous work studies multi-source UDA re-ID problem, we apply several general multi-source UDA methods to the state-of-the-art work MMT to compare with ours. It can be seen that our method is still superior to these comparison methods under the multi-source setting. In particular, when selecting the large scale dataset MSMT as target, our method outperforms them by 12+\% mAP and 15+\% top-1 accuracy. Moreover, in some tasks, our unsupervised results even \textbf{achieve comparable results with some popular fully-supervised methods} such as PCB \cite{PCB} and bag-of-tricks \cite{bagoftricks}. This result again verifies the effectiveness of our method. Note that no extra post-processing technique is adopted.

\subsection{Further Improvements on Source Domain}
Finally, we study the model performance on source-domain testsets after domain adaptation, and list related results in Table \ref{tab:test-on-source}, where `Pre-trained model' is obtained by supervised training on the corresponding single-source dataset. From the rows of `Pre-trained model' and `MMT (DBSCAN)', we can see that state-of-the-art UDA person re-ID methods usually forget the source-domain knowledge after ﬁne-tuning the pre-trained model on the target domain. Employing source data into domain adaptation fine-tuning stage is much better as shown in `MMT (3 Sources + 1 target)'. In comparison, the proposed method adapt to multiple domains more effectively, which further improves the result. Our domain adaption model is obviously superior to pre-trained model, and even outperforms popular supervised re-ID methods \cite{PCB,bagoftricks}. Such a phenomenon indicates that our method could also be applied to improve the supervised training.

\section{Discussion about the Module Compatibility}
In aforementioned experiments, RDSBN and MDIF modules show good compatibility. We consider that they complement each other. Specifically, on one hand, RDSBN module cannot completely eliminate domain-specific information, MDIF helps to further reduce domain gaps by fusing features. On the other hand, directly applying MDIF module is prone to incur noisy features since the original domain gaps are very large. RDSBN module may just establish a favourable condition for MDIF module.

\section{Conclusion}
In this work, we propose a multi-source framework for the unsupervised domain adaptive person re-ID task. Under the multi-source setting, simply combining different datasets brings limited improvement due to domain gaps. We propose to alleviate this problem from domain-specific view and domain-fusion view. Correspondingly, two constructive modules are developed, \ie{} RDSBN module and MDIF module. The former can reduce domain-specific characteristics and enhance distinctiveness of person features. The latter explores domain-agent-nodes to represent global domain information and uses a GCN structure to fuse features from different domains. Extensive experiments demonstrate that the proposed method outperforms state-of-the-art UDA person re-ID methods by a large margin, and even achieve comparable results to some popular fully supervised works.

\appendix

\begin{appendices}
\section{Details about Datasets}
We evaluate the proposed method on four widely-used person re-ID datasets: Market1501\cite{market1501}, DukeMTMC-reID\cite{ristani2016MTMC}, CUHK03\cite{cuhk03} and MSMT\cite{msmt}. 

\textbf{Market1501} contains 32,217 images of 1,501 identities captured by six cameras, where 12,936 images of 751 identities are for training, the remaining 19,281 images of 750 identities are for testing. The test set is further divided into a query
set consisting of 3,368 images and a gallery set consisting  of 15,913 images.

\textbf{DukeMTMC-reID} is collected in campus with eight cameras, which contains 36,411 images of 1,404 identities in total. Each identity is captured by at least two cameras.
The train set contains 16,522 images of 702 identities. The rest data are for testing where the query set contains 2,228 images and the gallery set contains 17,661 images. 

\textbf{CUHK03} contains 28,193 images of 1,467 identities captured by two cameras. 26,263 images of 1,367 identities are used for training. The query set and gallery set have 200 and 1,730 images of the remaining 100 identities, respectively.

\textbf{MSMT} consists of 126,441 images of 4,101 identities captured by 15 cameras. The train set contains 32,621 images of 1,041 identities. The remaining images of 3,060 identities are assigned to query set (11,659 images) and gallery set (82,161 images).

\section{Details of MDIF inference}
\begin{figure}
    \centering
    \includegraphics[width=0.93\linewidth]{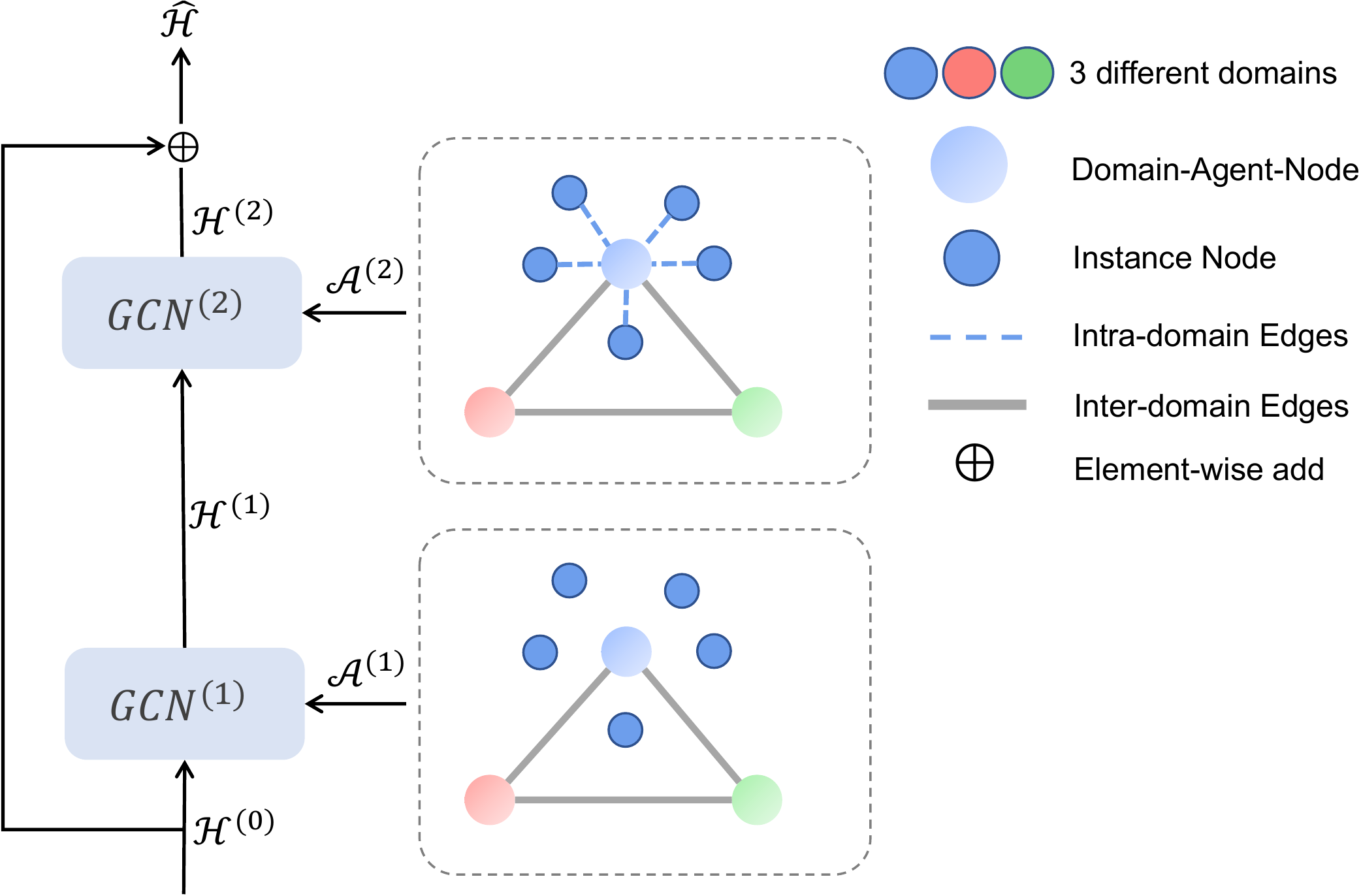}
    \caption{The illustration of GCN-based MDIF module in inference mode. For simplicity, we present two source domains (red and green) and one target domain (blue) here.}
    \label{fig:gcn_infer}
    \vspace{-10pt}
\end{figure}

Fig.~\ref{fig:gcn_infer} shows the graph structure of MDIF module in inference mode. Features are calculated according to this structure, and we fetch the $\hat{\mathcal{H}}$ as the fused features. With respect to the graph information, (1) as explained in Section 3.3.1 in our paper, we use the recorded moving average values of domain-agent-nodes in inference. Therefore, domain-agent-nodes will not be affected by target samples, and edges among domain-agent-nodes are also fixed. (2) While please note that the whole graph structure is not necessary to be fixed. As for $Q$, \textit{i.e.} the batch size in inference, it can be set to any number within the limitation of hard devices. The adjacency matrix $\mathcal{A}^{(l)}$ depicts connection relationships among graph nodes, it can be automatically tuned according to $Q$. In addition, the feature of a certain image will not be affected by other images since there is no connection between two instance nodes.

\section{More Ablation Studies}
To comprehensively validate the effectiveness and robustness of the proposed method, more ablation study experiments under another source-target setting are conducted and shown in this section. In accordance with previous experiments, two source datasets and one target dataset are employed for ablation study.

\subsection{Effectiveness of RDSBN}
\label{RDSBN_exp_supp}
As can be seen from Table \ref{tab:rdsbn_supp}, the observations are consistent with previous results, \ie the proposed RDSBN module is obviously superior to other normalization methods. Specifically, our method outperforms DSBN \cite{dsbn}, a recent and powerful domain adaptation work, by 3.1\% mAP and 1.1\% rank-1 accuracy. This again verifies the effectiveness of our RDSBN module.

\begin{table}[htb]
	\centering
	\caption{Ablation studies on RDSBN module under another source-target setting.}
	\vspace{5pt}
	\begin{tabular}{P{4.0cm}|C{0.55cm}C{0.55cm}C{0.55cm}C{0.73cm}}
 	\Xhline{2\arrayrulewidth}
	\multicolumn{1}{c|}{\multirow{2}{*}{Methods}} &  \multicolumn{4}{c}{Market+CUHK$\to$Duke} \\
	\cline{2-5}
	\multicolumn{1}{c|}{} & mAP & R-1 & R-5 & R-10\\ 
    % \hline
    \Xhline{2\arrayrulewidth}
    ResNet50-\textit{BN} (baseline) &  58.3 & 72.8 & 84.4 & 88.2 \\
    ResNet50-\textit{BIN} &  59.2 & 73.1 & 84.1 & 88.0 \\
    ResNet50-\textit{IBN} &  61.9 & 75.9 & 87.6 & 90.6 \\
    ResNet50-\textit{SN} &  56.7 & 73.5 & 84.0 & 87.7 \\
    ResNet50-\textit{DSBN} &  60.4 & 76.2 & 86.1 & 90.1  \\
    ResNet50-\textit{RDSBN} & \textbf{63.5} & \textbf{77.3} & \textbf{88.3} & \textbf{91.6}  \\
    \Xhline{2\arrayrulewidth}
	\end{tabular}
\label{tab:rdsbn_supp}
\end{table}

\begin{table*}[ht]
	\centering
	\caption{Analysis about inter-class distance and intra-class variance.}
	\vspace{5pt}
	\begin{tabular}{P{4cm}|C{1.2cm}C{1.2cm}C{1.2cm}C{3cm}}
 	\Xhline{2\arrayrulewidth}
 	Metrics \& Methods & Market & Duke & CUHK03 & Combined Dataset\\
    \Xhline{2\arrayrulewidth}
    inter-class distance \textit{DSBN} & 1.80 & 1.83 & 1.84 & 1.92\\
    inter-class distance \textit{RDSBN} & \textbf{1.88} & \textbf{1.85} & \textbf{1.91} & \textbf{1.95}\\
    \hline
    intra-class variance \textit{DSBN} & 0.35 & 0.36 & 0.32 & 0.34\\
    intra-class variance \textit{RDSBN} & \textbf{0.32} & \textbf{0.33} & \textbf{0.30} & \textbf{0.32}\\
    \Xhline{2\arrayrulewidth}
	\end{tabular}
\label{tab:rdsbn_feature_dist_supp}
\end{table*}

Practically, the feature distance is usually used to measure whether two samples belong to the same identity or not. Thus we further validate the effect of RDSBN module by analyzing inter-class and intra-class feature distances. The ability of \textit{pushing inter-class samples away} and \textit{pulling intra-class samples close} is a widely-used criterion in representation learning \cite{zhuang2019local, huang2019unsupervised}. To prove the \textit{`pushing away'} ability, we use the mean feature of samples to represent the corresponding identity, and calculate inter-class distance by averaging the pair-wise distances of all persons. The calculation can be formulated as follows. 
\begin{equation}
    \overline{D} = \frac{1}{M} \sum_{i,j \in Y}d(i,j),
\end{equation}
where $\overline{D}$ denotes the inter-class distance, $d(i,j)$ represents the Euclidean feature distance between identity $i$ and $j$, $Y$ is the label set and $M$ is the number of pair-wise identity distances.
A larger inter-class distance indicates that the \textit{`pushing away'} ability is stronger and identities are easier to distinguish. To prove the \textit{`pulling close'} ability, we first calculate the feature variance of each identity, then measure the intra-class variance by averaging these variances. This process can be formulated by
\begin{equation}
    \sigma_i^2 = \sum_{k=1}^K{(X_{i, k} - \mu_i)}^T(X_{i, k} - \mu_i),
\end{equation}
\begin{equation}
    \overline{\sigma^2} = \frac{1}{N}\sum_{i=1}^N \sigma_i^2,
\end{equation}
where $X_{i, k}$, $\sigma_i^2$, $\mu_i$ and $K$ denote the \textit{k}-th feature, feature variance, mean feature and sample number of identity $i$, $\overline{\sigma^2}$ denotes the intra-class variance, $N$ is the number of identities here.
A smaller intra-class variance indicates that the \textit{`pulling close'} ability is stronger and intra-class samples are more compact in feature space.

The corresponding experimental results are shown in Table \ref{tab:rdsbn_feature_dist_supp}, in which the Combined Dataset denotes the combination of Market, Duke and CUHK03. Compared to DSBN, the proposed RDSBN module achieves larger inter-class distance and smaller intra-class variance in all datasets. This result demonstrates that our RDSBN module can extract better features for re-ID task. Note that MDIF module is not used in this experiment.

\subsection{Effectiveness of MDIF.}
Performances of the proposed MDIF module under different model settings are reported in Table \ref{tab:mdif_ablation_supp}. We can observe that placing MDIF module on different base models always boosts performance significantly. Applying MDIF module to baseline achieves 4.4\% mAP and 4.1\% rank-1 accuracy improvement. Even building upon our RDSBN module, MDIF also brings 1.5\% mAP and 1.8\% rank-1 accuracy improvement.

The effect of domain-agent-node is also validated in Market+CUHK$\to$Duke task. As shown in Table \ref{tab:mdif_design_supp}, the proposed domain-agent-node, especially `Weighted Mean Agent Node', outperforms other variants stably. In addition, the inferior result of `intra-connections in $A^{(1)}$' verifies again that it is reasonable to omit intra-connections in the first GCN layer of MDIF module.

\begin{table}[h]
	\centering
	\caption{Ablation studies on MDIF module under another source-target setting.}
	\vspace{5pt}
	\begin{tabular}{P{4.0cm}|C{0.55cm}C{0.55cm}C{0.55cm}C{0.73cm}}
 	\Xhline{2\arrayrulewidth}
	\multicolumn{1}{c|}{\multirow{2}{*}{Methods}} &  \multicolumn{4}{c}{Market+CUHK$\to$Duke} \\
	\cline{2-5}
	\multicolumn{1}{c|}{} & mAP & R-1 & R-5 & R-10\\ 
% 	\hline \hline
    \Xhline{2\arrayrulewidth}
    ResNet50-\textit{BN} (baseline) & 58.3 & 72.8 & 84.4 & 88.2 \\
    ResNet50-\textit{BN-MDIF} & \textbf{62.7} & \textbf{76.9} & \textbf{87.8} & \textbf{91.2}  \\
    \hline
    ResNet50-\textit{DSBN} & 60.4 & 76.2 & 86.1 & 90.1  \\
    ResNet50-\textit{DSBN-MDIF} & \textbf{63.3} & \textbf{78.1} & \textbf{88.2} & \textbf{91.5}  \\
    \hline
    ResNet50-\textit{RDSBN} & 63.5 & 77.3 & 88.3 & 91.6  \\
    ResNet50-\textit{RDSBN-MDIF} & \textbf{65.0} & \textbf{79.1} & \textbf{88.5} & \textbf{91.6}  \\
    \Xhline{2\arrayrulewidth}
	\end{tabular}
\label{tab:mdif_ablation_supp}
\end{table}

\begin{table}[h]
	\centering
	\caption{Analysis of MDIF module design under another source-target setting.}
	\vspace{5pt}
	\begin{tabular}{P{4.0cm}|C{0.55cm}C{0.55cm}C{0.55cm}C{0.73cm}}
 	\Xhline{2\arrayrulewidth}
	\multicolumn{1}{c|}{\multirow{2}{*}{Methods}} & \multicolumn{4}{c}{Market+CUHK$\to$Duke}\\
	\cline{2-5}
	\multicolumn{1}{c|}{} &mAP & R-1 & R-5 & R-10\\ 
% 	\hline \hline
    \Xhline{2\arrayrulewidth}
    baseline & 58.3 & 72.8 & 84.4 & 88.2 \\
    \hline
    Instance Graph & 61.6 & 76.1 & 86.8 & 90.0 \\
    Mean Agent Node & 62.5 & 76.9 & 87.0 & 90.9 \\
    Weighted Mean Agent Node & \textbf{62.7} & \textbf{77.3} & \textbf{87.8} & \textbf{91.2} \\
    intra-connections in $A^{(1)}$ & 61.9 & 76.0 & 87.0 & 90.2 \\
    \Xhline{2\arrayrulewidth}
	\end{tabular}
\label{tab:mdif_design_supp}
\end{table}

\begin{table*}[htb]
	\centering
	\caption{Experiments on different domain adaptation frameworks.}
	\begin{tabular}{P{3.4cm}|C{1.3cm}C{0.55cm}C{0.55cm}C{0.73cm}|C{1.3cm}C{0.55cm}C{0.55cm}C{0.73cm}}
	\Xhline{2\arrayrulewidth}
	\multicolumn{1}{c|}{\multirow{2}{*}{Methods}} & \multicolumn{4}{c|}{Duke+CUHK$\to$Market} &  \multicolumn{4}{c}{Market+CUHK$\to$Duke}\\
	\cline{2-9}
	\multicolumn{1}{c|}{} & mAP & R-1 & R-5 & R-10 & mAP & R-1 & R-5 & R-10\\ 
    \Xhline{2\arrayrulewidth}
    Baseline & 70.1 & 86.7 & 95.0 & 96.5 & 58.3 & 72.8 & 84.4 & 88.2 \\
    Baseline+\textit{DSBN} & 74.1 & 89.2 & 96.2 & 97.9 & 60.4 & 76.2 & 86.1 & 90.1  \\
    Baseline+\textit{Ours} & \textbf{79.4} & \textbf{92.1} & \textbf{97.1} & \textbf{98.1} & \textbf{65.0} & \textbf{79.1} & \textbf{88.5} & \textbf{91.6}  \\
    \hline
    MMT & 75.3 & 89.6 & 96.5 & 97.6 & 62.6 & 76.2 & 87.1 & 90.2 \\
    MMT+\textit{DSBN} & 81.2 & 92.7 & 97.5 & 98.4 & 65.7 & 79.0 & 89.0 & 92.1 \\
    MMT+\textit{Ours} & \textbf{85.2} & \textbf{94.2} & \textbf{98.0} & \textbf{98.8} & \textbf{69.0} & \textbf{81.2} & \textbf{90.3} & \textbf{93.2} \\
    \Xhline{2\arrayrulewidth}
	\end{tabular}
\label{tab:basemodel_supp}
\end{table*}

\subsection{Effectiveness of the Proposed Method on Different Base Frameworks}
To further confirm the effectiveness and robustness of the proposed method, we apply it to the ordinary baseline framework and the state-of-the-art MMT \cite{mmt} framework, respectively. Results are shown in Table \ref{tab:basemodel_supp}. We can see that our method achieves 9.3\% mAP and 5.4\% rank-1 accuracy improvement on baseline, and 9.9\% mAP and 4.6\% rank-1 accuracy improvement on MMT. The proposed method also outperforms their variants equipped with DSBN . These stable improvements demonstrate the superiority of our method.
\end{appendices}

{\small
\bibliographystyle{ieee_fullname}
\bibliography{cvpr}

\begin{thebibliography}{10}\itemsep=-1pt

\bibitem{DSN}
Konstantinos Bousmalis, George Trigeorgis, Nathan Silberman, Dilip Krishnan,
  and Dumitru Erhan.
\newblock Domain separation networks.
\newblock In {\em Advances in Neural Information Processing Systems 29: Annual
  Conference on Neural Information Processing Systems 2016, December 5-10,
  2016, Barcelona, Spain}, pages 343--351, 2016.

\bibitem{dsbn}
Woong{-}Gi Chang, Tackgeun You, Seonguk Seo, Suha Kwak, and Bohyung Han.
\newblock Domain-specific batch normalization for unsupervised domain
  adaptation.
\newblock In {\em {IEEE} Conference on Computer Vision and Pattern Recognition,
  {CVPR} 2019, Long Beach, CA, USA, June 16-20, 2019}, pages 7354--7362, 2019.

\bibitem{spgan}
Weijian Deng, Liang Zheng, Qixiang Ye, Guoliang Kang, Yi Yang, and Jianbin
  Jiao.
\newblock Image-image domain adaptation with preserved self-similarity and
  domain-dissimilarity for person re-identification.
\newblock In {\em CVPR}, 2018.

\bibitem{espgan}
Weijian Deng, Liang Zheng, Qixiang Ye, Yi Yang, and Jianbin Jiao.
\newblock Similarity-preserving image-image domain adaptation for person
  re-identification.
\newblock {\em CoRR}, abs/1811.10551, 2018.

\bibitem{dbscan}
Martin Ester, Hans{-}Peter Kriegel, J{\"{o}}rg Sander, and Xiaowei Xu.
\newblock A density-based algorithm for discovering clusters in large spatial
  databases with noise.
\newblock In {\em Proceedings of the Second International Conference on
  Knowledge Discovery and Data Mining (KDD-96), Portland, Oregon, {USA}}, pages
  226--231, 1996.

\bibitem{pul}
Hehe Fan, Liang Zheng, Chenggang Yan, and Yi Yang.
\newblock Unsupervised person re-identification: Clustering and fine-tuning.
\newblock {\em {ACM} Trans. Multim. Comput. Commun. Appl.}, 14(4):83:1--83:18,
  2018.

\bibitem{ssg}
Yang Fu, Yunchao Wei, Guanshuo Wang, Yuqian Zhou, Honghui Shi, and Thomas~S.
  Huang.
\newblock Self-similarity grouping: A simple unsupervised cross domain
  adaptation approach for person re-identification.
\newblock In {\em The IEEE International Conference on Computer Vision (ICCV)},
  October 2019.

\bibitem{grl}
Yaroslav Ganin and Victor Lempitsky.
\newblock Unsupervised domain adaptation by backpropagation.
\newblock In {\em International conference on machine learning}, pages
  1180--1189. PMLR, 2015.

\bibitem{UDAB}
Yaroslav Ganin and Victor~S. Lempitsky.
\newblock Unsupervised domain adaptation by backpropagation.
\newblock In {\em Proceedings of the 32nd International Conference on Machine
  Learning, {ICML} 2015, Lille, France, 6-11 July 2015}, volume~37, pages
  1180--1189, 2015.

\bibitem{mmt}
Yixiao Ge, Dapeng Chen, and Hongsheng Li.
\newblock Mutual mean-teaching: Pseudo label refinery for unsupervised domain
  adaptation on person re-identification.
\newblock In {\em International Conference on Learning Representations}, 2020.

\bibitem{resnet}
Kaiming He, Xiangyu Zhang, Shaoqing Ren, and Jian Sun.
\newblock Deep residual learning for image recognition.
\newblock In {\em 2016 {IEEE} Conference on Computer Vision and Pattern
  Recognition, {CVPR} 2016, Las Vegas, NV, USA, June 27-30, 2016}, pages
  770--778, 2016.

\bibitem{triplet}
Alexander Hermans*, Lucas Beyer*, and Bastian Leibe.
\newblock {In Defense of the Triplet Loss for Person Re-Identification}.
\newblock {\em arXiv preprint arXiv:1703.07737}, 2017.

\bibitem{huang2019unsupervised}
Jiabo Huang, Qi Dong, Shaogang Gong, and Xiatian Zhu.
\newblock Unsupervised deep learning by neighbourhood discovery.
\newblock {\em arXiv preprint arXiv:1904.11567}, 2019.

\bibitem{adain}
Xun Huang and Serge Belongie.
\newblock Arbitrary style transfer in real-time with adaptive instance
  normalization.
\newblock In {\em ICCV}, 2017.

\bibitem{phgcn}
Bo Jiang, Xixi Wang, and Bin Luo.
\newblock {PH-GCN:} person re-identification with part-based hierarchical graph
  convolutional network.
\newblock {\em CoRR}, 2019.

\bibitem{jin2020style}
Xin Jin, Cuiling Lan, Wenjun Zeng, Zhibo Chen, and Li Zhang.
\newblock Style normalization and restitution for generalizable person
  re-identification.
\newblock {\em CVPR}, 2020.

\bibitem{adam}
Diederik~P. Kingma and Jimmy Ba.
\newblock Adam: {A} method for stochastic optimization.
\newblock In {\em 3rd International Conference on Learning Representations,
  {ICLR} 2015, San Diego, CA, USA, May 7-9, 2015, Conference Track
  Proceedings}, 2015.

\bibitem{gcn}
Thomas~N. Kipf and Max Welling.
\newblock Semi-supervised classification with graph convolutional networks.
\newblock In {\em International Conference on Learning Representations (ICLR)},
  2017.

\bibitem{cuhk03}
W. {Li}, R. {Zhao}, T. {Xiao}, and X. {Wang}.
\newblock Deepreid: Deep filter pairing neural network for person
  re-identification.
\newblock In {\em 2014 IEEE Conference on Computer Vision and Pattern
  Recognition}, pages 152--159, 2014.

\bibitem{lin2020corsscam}
Yutian Lin, Yu Wu, Chenggang Yan, Mingliang Xu, and Yi Yang.
\newblock Unsupervised person re-identification via cross-camera similarity
  exploration.
\newblock {\em IEEE Transactions on Image Processing}, 29:5481--5490, 2020.

\bibitem{lin2020soften}
Yutian Lin, Lingxi Xie, Yu Wu, Chenggang Yan, and Qi Tian.
\newblock Unsupervised person re-identification via softened similarity
  learning.
\newblock In {\em Proceedings of the IEEE/CVF Conference on Computer Vision and
  Pattern Recognition}, pages 3390--3399, 2020.

\bibitem{DAN}
Mingsheng Long, Yue Cao, Jianmin Wang, and Michael Jordan.
\newblock Learning transferable features with deep adaptation networks.
\newblock In {\em International conference on machine learning}, pages 97--105.
  PMLR, 2015.

\bibitem{RTN}
Mingsheng Long, Han Zhu, Jianmin Wang, and Michael~I Jordan.
\newblock Unsupervised domain adaptation with residual transfer networks.
\newblock In {\em Advances in neural information processing systems}, pages
  136--144, 2016.

\bibitem{JAN}
Mingsheng Long, Han Zhu, Jianmin Wang, and Michael~I Jordan.
\newblock Deep transfer learning with joint adaptation networks.
\newblock In {\em International conference on machine learning}, pages
  2208--2217. PMLR, 2017.

\bibitem{bagoftricks}
Hao Luo, Youzhi Gu, Xingyu Liao, Shenqi Lai, and Wei Jiang.
\newblock Bag of tricks and a strong baseline for deep person
  re-identification.
\newblock In {\em Proceedings of the IEEE Conference on Computer Vision and
  Pattern Recognition Workshops}, pages 0--0, 2019.

\bibitem{SN}
Ping Luo, Jiamin Ren, Zhanglin Peng, Ruimao Zhang, and Jingyu Li.
\newblock Differentiable learning-to-normalize via switchable normalization.
\newblock {\em International Conference on Learning Representation (ICLR)},
  2019.

\bibitem{reco}
Chen Ma, Liheng Ma, Yingxue Zhang, Jianing Sun, Xue Liu, and Mark Coates.
\newblock Memory augmented graph neural networks for sequential recommendation.
\newblock In {\em The Thirty-Fourth {AAAI} Conference on Artificial
  Intelligence}, pages 5045--5052, 2020.

\bibitem{BIN}
Hyeonseob Nam and Hyo-Eun Kim.
\newblock Batch-instance normalization for adaptively style-invariant neural
  networks.
\newblock In {\em Advances in Neural Information Processing Systems}, 2018.

\bibitem{IBN-Net}
Xingang Pan, Ping Luo, Jianping Shi, and Xiaoou Tang.
\newblock Two at once: Enhancing learning and generalization capacities via
  ibn-net.
\newblock In {\em ECCV}, 2018.

\bibitem{MMMSDA}
Xingchao Peng, Qinxun Bai, Xide Xia, Zijun Huang, Kate Saenko, and Bo Wang.
\newblock Moment matching for multi-source domain adaptation.
\newblock In {\em 2019 {IEEE/CVF} International Conference on Computer Vision,
  {ICCV} 2019, Seoul, Korea (South), October 27 - November 2, 2019}, pages
  1406--1415, 2019.

\bibitem{peng2019moment}
Xingchao Peng, Qinxun Bai, Xide Xia, Zijun Huang, Kate Saenko, and Bo Wang.
\newblock Moment matching for multi-source domain adaptation.
\newblock In {\em Proceedings of the IEEE International Conference on Computer
  Vision}, pages 1406--1415, 2019.

\bibitem{ristani2016MTMC}
Ergys Ristani, Francesco Solera, Roger Zou, Rita Cucchiara, and Carlo Tomasi.
\newblock Performance measures and a data set for multi-target, multi-camera
  tracking.
\newblock In {\em European Conference on Computer Vision workshop on
  Benchmarking Multi-Target Tracking}, 2016.

\bibitem{sggnn}
Yantao Shen, Hongsheng Li, Shuai Yi, Dapeng Chen, and Xiaogang Wang.
\newblock Person re-identification with deep similarity-guided graph neural
  network.
\newblock In {\em Computer Vision - {ECCV} 2018 - 15th European Conference,
  Munich, Germany, September 8-14, 2018, Proceedings, Part {XV}}, volume 11219,
  pages 508--526, 2018.

\bibitem{udatp}
Liangchen Song, Cheng Wang, Lefei Zhang, Bo Du, Qian Zhang, Chang Huang, and
  Xinggang Wang.
\newblock Unsupervised domain adaptive re-identification: Theory and practice.
\newblock {\em Pattern Recognition}, 102:107173, 2020.

\bibitem{PCB}
Yifan Sun, Liang Zheng, Yi Yang, Qi Tian, and Shengjin Wang.
\newblock Beyond part models: Person retrieval with refined part pooling (and
  {A} strong convolutional baseline).
\newblock In {\em Computer Vision - {ECCV} 2018 - 15th European Conference,
  Munich, Germany, September 8-14, 2018, Proceedings, Part {IV}}, volume 11208,
  pages 501--518, 2018.

\bibitem{gat}
Petar Veli{\v{c}}kovi{\'{c}}, Guillem Cucurull, Arantxa Casanova, Adriana
  Romero, Pietro Li{\`{o}}, and Yoshua Bengio.
\newblock {Graph Attention Networks}.
\newblock {\em International Conference on Learning Representations}, 2018.

\bibitem{wang2020unsupervised}
Dongkai Wang and Shiliang Zhang.
\newblock Unsupervised person re-identification via multi-label classification.
\newblock In {\em CVPR}, 2020.

\bibitem{MGN}
Guanshuo Wang, Yufeng Yuan, Xiong Chen, Jiwei Li, and Xi Zhou.
\newblock Learning discriminative features with multiple granularities for
  person re-identification.
\newblock In {\em 2018 {ACM} Multimedia Conference on Multimedia Conference,
  {MM} 2018, Seoul, Republic of Korea, October 22-26, 2018}, pages 274--282,
  2018.

\bibitem{msmt}
L. {Wei}, S. {Zhang}, W. {Gao}, and Q. {Tian}.
\newblock Person transfer gan to bridge domain gap for person
  re-identification.
\newblock In {\em 2018 IEEE/CVF Conference on Computer Vision and Pattern
  Recognition}, pages 79--88, 2018.

\bibitem{leakyrelu}
Bing Xu, Naiyan Wang, Tianqi Chen, and Mu Li.
\newblock Empirical evaluation of rectified activations in convolutional
  network.
\newblock {\em arXiv preprint arXiv:1505.00853}, 2015.

\bibitem{YanXL18}
Sijie Yan, Yuanjun Xiong, and Dahua Lin.
\newblock Spatial temporal graph convolutional networks for skeleton-based
  action recognition.
\newblock In {\em Proceedings of the Thirty-Second {AAAI} Conference on
  Artificial Intelligence}, pages 7444--7452, 2018.

\bibitem{asymmetric}
Fengxiang Yang, Ke Li, Zhun Zhong, Zhiming Luo, Xing Sun, Hao Cheng, Xiaowei
  Guo, Feiyue Huang, Rongrong Ji, and Shaozi Li.
\newblock Asymmetric co-teaching for unsupervised cross-domain person
  re-identification.
\newblock In {\em AAAI}, pages 12597--12604, 2020.

\bibitem{CMSS}
Luyu Yang, Yogesh Balaji, Ser{-}Nam Lim, and Abhinav Shrivastava.
\newblock Curriculum manager for source selection in multi-source domain
  adaptation.
\newblock {\em CoRR}, abs/2007.01261, 2020.

\bibitem{CMD}
Werner Zellinger, Thomas Grubinger, Edwin Lughofer, Thomas Natschl{\"a}ger, and
  Susanne Saminger-Platz.
\newblock Central moment discrepancy (cmd) for domain-invariant representation
  learning.
\newblock {\em arXiv preprint arXiv:1702.08811}, 2017.

\bibitem{adcluster}
Yunpeng Zhai, Shijian Lu, Qixiang Ye, Xuebo Shan, Jie Chen, Rongrong Ji, and
  Yonghong Tian.
\newblock Ad-cluster: Augmented discriminative clustering for domain adaptive
  person re-identification.
\newblock In {\em 2020 {IEEE/CVF} Conference on Computer Vision and Pattern
  Recognition, {CVPR} 2020, Seattle, WA, USA, June 13-19, 2020}, pages
  9018--9027, 2020.

\bibitem{CAN}
Weichen Zhang, Wanli Ouyang, Wen Li, and Dong Xu.
\newblock Collaborative and adversarial network for unsupervised domain
  adaptation.
\newblock In {\em Proceedings of the IEEE Conference on Computer Vision and
  Pattern Recognition}, pages 3801--3809, 2018.

\bibitem{past}
Xinyu Zhang, Jiewei Cao, Chunhua Shen, and Mingyu You.
\newblock Self-training with progressive augmentation for unsupervised
  cross-domain person re-identification.
\newblock In {\em Proceedings of the IEEE International Conference on Computer
  Vision}, pages 8222--8231, 2019.

\bibitem{RGA}
Zhizheng Zhang, Cuiling Lan, Wenjun Zeng, Xin Jin, and Zhibo Chen.
\newblock Relation-aware global attention for person re-identification.
\newblock In {\em Proceedings of the IEEE/CVF Conference on Computer Vision and
  Pattern Recognition (CVPR)}, June 2020.

\bibitem{nrmt}
Fang Zhao, Shengcai Liao, Guo-Sen Xie, Jian Zhao, Kaihao Zhang, and Ling Shao.
\newblock Unsupervised domain adaptation with noise resistible mutual-training
  for person re-identification.
\newblock In {\em European Conference on Computer Vision}, pages 526--544.
  Springer, 2020.

\bibitem{market1501}
Liang Zheng, Liyue Shen, Lu Tian, Shengjin Wang, Jingdong Wang, and Qi Tian.
\newblock Scalable person re-identification: A benchmark.
\newblock In {\em Proceedings of the 2015 IEEE International Conference on
  Computer Vision (ICCV)}, ICCV '15, page 1116–1124, 2015.

\bibitem{adgcn}
Jia-Xing Zhong, Nannan Li, Weijie Kong, Shan Liu, Thomas~H. Li, and Ge Li.
\newblock Graph convolutional label noise cleaner: Train a plug-and-play action
  classifier for anomaly detection.
\newblock In {\em The IEEE Conference on Computer Vision and Pattern
  Recognition (CVPR)}, 2019.

\bibitem{zhong2020learning}
Zhun Zhong, Liang Zheng, Zhiming Luo, Shaozi Li, and Yi Yang.
\newblock Learning to adapt invariance in memory for person re-identification.
\newblock {\em TPAMI}, 2020.

\bibitem{zhuang2019local}
Chengxu Zhuang, Alex~Lin Zhai, and Daniel Yamins.
\newblock Local aggregation for unsupervised learning of visual embeddings.
\newblock In {\em Proceedings of the IEEE International Conference on Computer
  Vision}, pages 6002--6012, 2019.

\bibitem{dgnet_plus}
Yang Zou, Xiaodong Yang, Zhiding Yu, B.~V. K.~Vijaya Kumar, and Jan Kautz.
\newblock Joint disentangling and adaptation for cross-domain person
  re-identification.
\newblock {\em CoRR}, abs/2007.10315, 2020.

\end{thebibliography}
}

\end{document}